\documentclass[letterpaper, 10 pt, conference]{ieeeconf}

\IEEEoverridecommandlockouts                              %

\usepackage{amsmath}
\usepackage{amssymb}
\usepackage{url}
\usepackage{xcolor}
\usepackage{graphicx}
\graphicspath{{figures/}}
\usepackage{cite}     %
\usepackage[caption=false,font=footnotesize]{subfig}

\usepackage{todonotes}

\usepackage[breaklinks,colorlinks,citecolor=blue]{hyperref}

\usepackage{booktabs}
\usepackage{tikz}

\usepackage{amsthm}
\theoremstyle{definition}   %
\newtheorem{assumption}{Assumption}
\theoremstyle{definition}

\theoremstyle{plain}        %

\newtheorem{claim}{Claim}

\newcommand{\cost}{\ell}   %

\makeatletter
\let\clark@oldthebibliography\thebibliography
\renewcommand{\thebibliography}[1]{%
  \clark@oldthebibliography{#1}%
  \setlength{\itemsep}{0pt}%
  \setlength{\parsep}{0pt}%
  \setlength{\topsep}{2pt}}
\makeatother

\AtBeginDocument{
  \setlength{\textfloatsep}{11pt plus 2pt minus 2pt}
  \setlength{\floatsep}{9pt plus 2pt minus 2pt}
  \setlength{\intextsep}{9pt plus 2pt minus 2pt}
  \setlength{\abovecaptionskip}{4pt}
  \setlength{\belowcaptionskip}{0pt}
  \setlength{\abovedisplayskip}{6pt}
  \setlength{\belowdisplayskip}{6pt}
  \setlength{\abovedisplayshortskip}{6pt}
  \setlength{\belowdisplayshortskip}{6pt}
  \setlength{\jot}{4pt}
}

\title{\LARGE \bf
  Propagating Elevation-Map Uncertainty Through the Contact Maximum in
  Closed Form
}

\author{Ale\v{s} Ku\v{c}era$^{1}$ and Karel Zimmermann$^{1}$%
  \thanks{$^{1}$The authors are with the Department of Cybernetics, Faculty of
    Electrical Engineering, Czech Technical University in Prague, Prague, Czech Republic
    {\tt\small kuceral4@fel.cvut.cz}}%
}
\hypersetup{pdfauthor={Ales Kucera, Karel Zimmermann},
  pdftitle={Propagating Elevation-Map Uncertainty Through the Contact Maximum in Closed Form}}

\begin{document}
\raggedbottom

\maketitle
\thispagestyle{empty}
\pagestyle{empty}

\begin{abstract}
  Risk-aware planners score paths on uncertain elevation maps using the path
  cost's mean and standard deviation. Modeling rigid contact, however,
  requires computing a maximum over several uncertain cells. First-order
  propagation loses accuracy here by differentiating at only a single cell,
  while Monte Carlo sampling requires a full path evaluation per draw. We
  compute the moments of that contact maximum in closed form using Clark's
  pairwise recursion. By tracking each contact's covariance against the
  shared map cells, we propagate the smooth remainder using exact Gaussian
  quadratic-form identities. A contest-depth calibration, fitted once on two
  design traverses, closes the aggregate standard-deviation shortfall that
  remains. On $317$ held-out rover path segments, scored against a Monte
  Carlo reference from the same belief, every pre-registered criterion was
  met. The corrected Clark fold cuts the median error of the mean from
  linearization's $2.3\%$ to $0.19\%$ and attains the lowest
  $\mathrm{CVaR}_{0.9}$ error of every method tested. A benchmark plan costs
  just $5.5~\mu$s on a GPU. These accuracy gains concentrate at contested
  contacts. While they seldom change which path is chosen on this terrain,
  the fold still selects the reference-best path in $98\%$ of decisions
  against linearization's $93$ to $96\%$. On a second dataset the mean
  transfers, though the risk number does not.
\end{abstract}

\section{Introduction}
\label{sec:intro}

Robots navigating rough terrain rely on elevation maps subject to sensor
noise and estimation drift~\cite{fankhauser2018probabilistic}. To
ensure safety, risk-aware planners evaluate paths using metrics such as the
conditional value at risk (CVaR)~\cite{fan2021step}.
Assuming a Gaussian path cost, these metrics are fully dictated by the
cost's mean and standard deviation~\cite{fan2021step}.

\begin{figure}[t]
  \centering
  \includegraphics[width=\columnwidth]{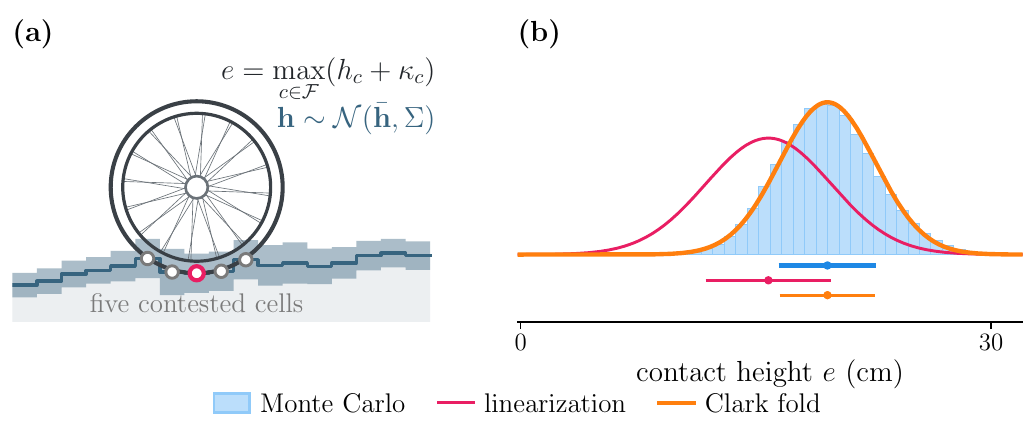}
  \caption{\textbf{(a)} A wheel over Gaussian terrain heights (per-cell mean
  and $\pm 2\sigma$ band) settles on the cell that maximizes its elevation,
  adjusted for wheel geometry. The contact height $e$ is thus a maximum over
  underlying cells, five of the seven here contesting it; the five are
  ringed, and in pink the one highest in the mean map, the only cell
  linearization considers. \textbf{(b)} The distribution of $e$ (a
  constructed example, its two standard-deviation ratios matched to the
  medians at this contest depth, Section~\ref{sec:gate}). The bars are
  $200{,}000$ draws; the curve is the Gaussian carrying their mean and
  standard deviation ($19.6 \pm 3.1$~cm). Linearization misses both,
  underestimating the mean by $3.8$~cm and overestimating the standard
  deviation by $29\%$ ($15.8 \pm 4.0$); the Clark fold reproduces both. The deficit the corrected fold repairs is not here but in
  the aggregation over a plan (Section~\ref{sec:estimator}).}
  \label{fig:teaser}
\end{figure}

The computational bottleneck in evaluating this cost is rigid contact.
Because a wheel rests on the terrain point that maximizes its elevation
(Fig.~\ref{fig:teaser}a), a path's cost is driven by the maximum of several
uncertain cells. Standard first-order propagation fails here because it
differentiates only at the cell that wins on the mean map. This single
choice ruins the estimation of both moments: it discards the statistical
inflation that Jensen's inequality applies to the mean of a maximum, and it
reads the variance from one cell as though the others were not
competing for contact. Consequently, planners needing accurate moments must
fall back on expensive sampling, which demands one full rollout per draw.

A natural workaround is to soften the contact, replacing the maximum with a
temperature-weighted average as differentiable simulators do. However, no fixed
temperature can. A softmax weight is strictly a function of the gap
between two cells' mean elevations. In reality, the probability that a cell carries
the load depends on that gap measured in units of the margin's own spread. If the means are held constant while uncertainty grows,
the true probability approaches one half, yet the temperature-weighted
output remains unchanged (Section~\ref{sec:softening}). Recovering the
contact maximum's moments requires the belief's second moment, not a
smoother maximum.

We compute those moments in closed form. Clark's
identities~\cite{clark1961greatest} yield the exact first two moments of the
maximum of two Gaussians. Applied pairwise across a footprint, tracking each
folded contact's covariance against the shared map cells, they return the
contact height's mean and its covariance with every other contact. An exact
Gaussian quadratic form turns these into the path cost's moments. While the
mean needs nothing further, the standard deviation requires a single
calibration fitted once on design data
(Section~\ref{sec:estimator}). We term this pipeline the \emph{corrected
Clark fold}.

We evaluated the corrected fold on the BASEPROD semi-desert rover
dataset~\cite{gerdes2024baseprod}, benchmarking against a $20{,}000$-sample
Monte Carlo reference drawn from the same map beliefs to isolate propagation
accuracy. Every pre-registered criterion was met. Across $317$ held-out
segments, the corrected fold achieved a median mean error of $0.19\%$
(compared to $2.3\%$ for linearization) and the lowest median CVaR error,
$2.7$ times below linearization's.
This accuracy advantage is earned primarily at contested folds
(Section~\ref{sec:gate}); on this terrain, safety threshold verdicts do not
separate either fold from standard linearization
(Section~\ref{sec:experiments}-E).

Our main contributions are: (1) a characterization of rigid contact in
contest depth: why first-order propagation fails at it, why
no softened contact repairs that, and the depth at which the two
constructions coincide
(Sections~\ref{sec:onehot}, \ref{sec:softening}, \ref{sec:gate}); (2) a
GPU-accelerated closed form for the contact maximum, with the contest-depth
calibration that closes its aggregate standard-deviation deficit: the
corrected Clark fold (Section~\ref{sec:method} onward);
and (3) a pre-registered evaluation on
held-out rover traverses alongside a zero-shot transfer check on a second
dataset (Section~\ref{sec:experiments}).

\section{Related Work}
\label{sec:related}

We categorize prior work by how it resolves the maximum of uncertain heights
at a contact.

\textbf{Ignoring the maximum.} Methods like STEP~\cite{fan2021step}
aggregate local risk from the map into a path-level CVaR. Because the robot
is modeled using surface normals and footprint plane fits rather than
discrete contact heights, the maximum is never actually formed.

\textbf{Bounding the maximum.} The ACE framework~\cite{otsu2020ace} and its
probabilistic successor, \mbox{p-ACE}~\cite{ghosh2018pace}, evaluate
clearance using deterministic kinematic bounds. Because the maximum relies
on these bounds rather than the belief distribution, they cannot
analytically propagate the map's mean or standard deviation through the
contact.

\textbf{Replacing the maximum with a heuristic.} For planetary landing,
Tomita et al.~\cite{tomita2023stochastic} compute exact lander slope moments
but approximate the worst-case point under the footprint by raising a
probability to an exponent fitted offline against sampled terrains.

\textbf{Sampling.} Monte Carlo approaches (e.g.,
ProTerrain~\cite{raja2025proterrain}) draw correlated terrains and simulate
them: exact per sample, but requiring one full physics simulation per draw.

\textbf{Smoothing the maximum.} Differentiable
simulators~\cite{newbury2024review} expose terrain derivatives of a
rollout~\cite{agishev2024monoforce,kucera2026ostrich};
propagating the map's variance through them is a first-order second-moment
(FOSM) linearization. To bypass pathological hard-contact
gradients~\cite{zhong2022correct,paulus2025diffmjx}, remedies average them
over sampled perturbations~\cite{lelidec2022randomized,suh2022bundled} or use
a softmax.
However, softmax weights rely solely on mean elevations, ignoring
the probability that an uncertain point actually supports the contact
(Section~\ref{sec:softening}).

\textbf{Moments of the maximum.} Clark~\cite{clark1961greatest} derived
exact formulas for the maximum of two correlated Gaussians, using a pairwise
recursion for larger sets. Block-based statistical static timing analysis (SSTA)
propagates circuit delay uncertainty through this
recursion~\cite{visweswariah2004first}, and advanced variants buy back what
the Gaussian fit loses with a third moment and a skew-normal
maximum~\cite{chopra2006skew}. We instead locate the deficit we measure
(Section~\ref{sec:experiments}-C) in the aggregation rather than the fold,
and close it with a calibration in contest depth
(Section~\ref{sec:estimator}). To our
knowledge, no prior robotics work applies Clark's recursion to a geometric
terrain contact maximum.

\section{Analytic Moments of the Contact Maximum}
\label{sec:validity}

We compute the contact maximum's moments in closed form, without sampling
or softening.

\subsection{From non-penetration to a maximum}

For a ground vehicle under sustained gravity, rigid contact reduces to a
heightmap. Under the complementarity formulation, bodies cannot
interpenetrate~\cite{andrews2022contact}. Because gravity keeps the vehicle
grounded, we assume exactly one vertical constraint remains active at each
candidate contact.
Consequently, the
3-D terrain simplifies to a single supporting surface per column, which is
the output format probabilistic mapping
publishes~\cite{fankhauser2018probabilistic}.

\begin{assumption}[Gaussian terrain belief]\label{as:belief}
The planner's input is a Gaussian belief over the terrain heights,
$\mathbf{h} \sim \mathcal{N}(\bar{\mathbf{h}}, \Sigma)$ on a 2.5-D grid.
Everything below depends on the map only through
$(\bar{\mathbf{h}}, \Sigma)$.
\end{assumption}

We resolve contact quasi-statically (rigid body and terrain at equilibrium
under gravity), reading each wheel's support from the cells under its
footprint, as clearance evaluation reads per-wheel bounds from each wheel
box~\cite{otsu2020ace}; whether it reproduces this vehicle's
attitude is not measured here (Section~\ref{sec:limitations}).
Under this
equilibrium, a rigid wheel spanning footprint cells $\mathcal{F}$ rests at
the support height
\begin{equation}
  e \;=\; \max_{c \, \in \, \mathcal{F}} \left( h_c + \kappa_c \right),
  \label{eq:envelope}
\end{equation}
where $\kappa_c \leq 0$ is the wheel's geometric shape offset at cell $c$,
zero under the axle and most negative at the rim. Non-penetration bounds the
wheel's lowest point from below, $z \geq h_c + \kappa_c$, with $-\kappa_c$
the height of the wheel's surface above its lowest point at that cell's
offset, and gravity selects the lowest admissible $z$, which makes
\eqref{eq:envelope} the lower boundary of the wheel's configuration-space
obstacle~\cite{lozanoperez1983spatial}.
The vehicle composes several of
these supports through an affine map, which the delta-method step
(Section~\ref{sec:delta}) passes exactly, so one support's derivation
carries to the plan.

\subsection{The derivative is one-hot}
\label{sec:onehot}

Because the support height is a maximum, its derivative is one-hot.
Wherever the maximizer $c^{\star}$ of \eqref{eq:envelope} is unique,
\begin{equation}
  \frac{\partial e}{\partial h_c}
  \;=\; \mathbf{1}\!\left[\, c = c^{\star} \right].
  \label{eq:onehot}
\end{equation}
Away from ties the maximum is affine in the winner alone. The
derivative is blind to runners-up (a cell one millimeter from taking over
and a cell a meter below produce the same zero), carries no curvature for a
second-order correction, and jumps by a whole unit the instant $c^{\star}$
stops winning.

We call the footprint's validity radius the margin
$e - \max_{c \neq c^{\star}} (h_c + \kappa_c)$, evaluated at the mean map
$\bar{\mathbf{h}}$, by which the winning candidate leads the best of the
rest. First-order propagation estimates the support's moments from the
derivative at that mean map: $\mathbb{E}[e] \approx e(\bar{\mathbf{h}})$,
the maximum of the means, and
$\mathrm{Var}[e] \approx \nabla e^{\top} \Sigma \nabla e$, which
by \eqref{eq:onehot} collapses to $\sigma_{c^{\star}}^2$, the winner's
variance alone (Fig.~\ref{fig:teaser}b). Under
Assumption~\ref{as:belief} the belief always puts some mass past that
radius, so neither estimate is ever exact. By Jensen's inequality the
expectation of a maximum exceeds the maximum of the expectations unless one
candidate almost surely dominates, and a variance read off one cell
describes only the piece the gradient was taken on.

How much of the belief's mass lies past the radius depends on the width of
the contest, the spread of the margin that would have to be closed. Two
candidates under one wheel (write $H_i = h_{c_i} + \kappa_{c_i}$, with means
$\bar h_i$, variances $\sigma_i^2$ and covariance $\sigma_{12}$) settle their
contest through their difference $D = H_2 - H_1$, Gaussian with mean
$\bar h_2 - \bar h_1$ and variance
\begin{equation}
  a^2 \;=\; \sigma_1^2 + \sigma_2^2 - 2\sigma_{12}.
  \label{eq:diffwidth}
\end{equation}
Cell $2$ carries the wheel with probability
$\Pr[D > 0] = \Phi(\Delta/a)$, $\Delta = \bar h_2 - \bar h_1$; $\Phi$ and
$\varphi$ denote the standard normal cumulative distribution and density
throughout. Spatial correlation enters this contest only through the
difference: it cancels in a difference exactly as it compounds in a sum.
Correlated candidates therefore yield a settled winner, independent ones a
contested maximum, where linearization fails. Measured in units of $a$,
the validity radius is this
lead $\Delta/a$, the contest depth the rest of the paper reports. On the
held-out windows of Section~\ref{sec:experiments}, its cost-weighted median
is $0.62$ under foresight and $0.77$ under hindsight, so the typical window
is decided inside one contest width, which is where both first-order errors above are largest.

\begin{figure}[t]
  \centering
  \includegraphics[width=\columnwidth]{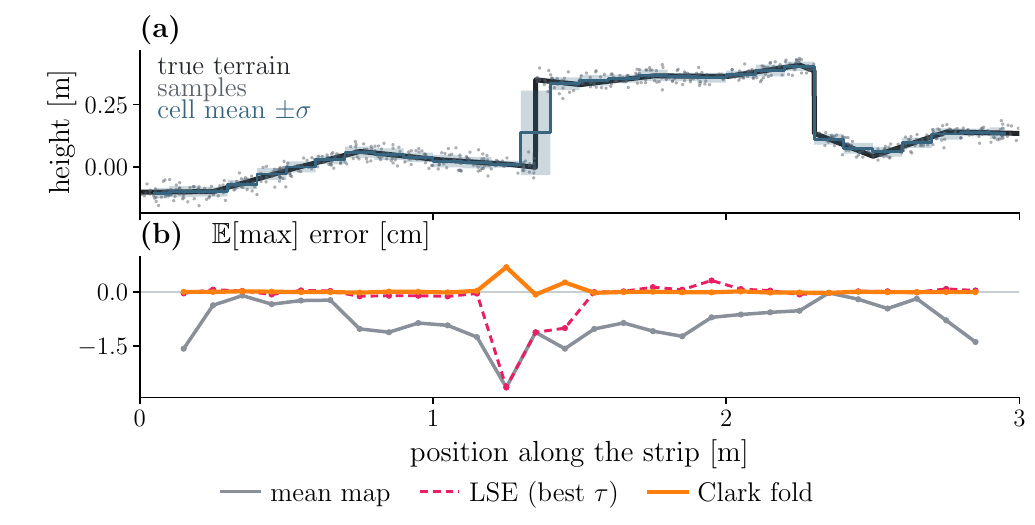}
  \caption{\textbf{(a)} A synthetic 1-D environment ($35$~cm step, $25$~cm
  drop) sampled with $\sigma = 2$~cm range noise ($25$ samples per $10$~cm
  cell) and rasterized into a Gaussian belief. At the step, samples straddle
  the discontinuity; the mean lands mid-air, and the standard deviation
  grows eightfold. \textbf{(b)} Error in $\mathbb{E}[\max]$ against a
  $20{,}000$-draw Monte Carlo reference for a flat $3$-cell footprint swept
  along the same strip, so the step in (a) stands over the cells it spoils.
  The mean map is systematically optimistic, and log-sum-exp
  at its best-fit temperature ($\tau = 1.6$~cm) returns the mean map's own
  value to within $0.002$~cm, its weight reading the gap between the means
  rather than the spread of the margin (Section~\ref{sec:softening}). RMS
  error: $0.14$~cm for the fold, $0.59$ for LSE, $1.00$ for the mean map.}
  \label{fig:toy1d}
\end{figure}

\subsection{Why softening the operator cannot recover it}
\label{sec:softening}

Replacing the maximum by a temperature-$\tau$ softmax changes the operator,
and it fails on the width of the contest (Fig.~\ref{fig:toy1d}b). The second
candidate's win probability is $\Phi(\Delta/a)$
(Section~\ref{sec:onehot}), while the softmax weight
$\mathrm{logistic}(\Delta/\tau)$ sees the mean difference $\Delta$ alone, and
no constant temperature repairs that. Hold the means fixed and grow the variances:
by \eqref{eq:diffwidth} the win probability tends to $\tfrac{1}{2}$, while a
function of the means cannot move. And because softmax weights are ordered
as the means, a cell known to lie below a certain leader, which can never
carry the wheel, still outranks an uncertain trailing cell that can.
Differentiate the expected maximum under the expectation instead, which is
licensed because the integrand is Lipschitz in $\bar h_i$ with derivative
bounded by one; that derivative is $\mathbf{1}[\,i = c^{\star}\,]$
by \eqref{eq:onehot}, whose expectation is the win probability, so for a
nondegenerate belief
\begin{equation}
  \frac{\partial\, \mathbb{E}[e]}{\partial \bar h_i}
  \;=\; \Pr[\,i \text{ carries the wheel}\,],
  \label{eq:smoothgrad}
\end{equation}
a bounded function, smooth in the means, that recovers \eqref{eq:onehot} as
the variances vanish. Equation~\eqref{eq:smoothgrad} is the
Williams--Daly--Zachary identity of random-utility theory, the gradient of
an expected maximum being the choice probabilities~\cite{sorensen2022mcfadden};
the quantity is the $\Phi(\alpha)$ of Clark's recursion,
\eqref{eq:clark-mean} below, and the tightness probability that timing
analysis reports for each input, and uses to blend the inputs' sensitivities
into the maximum's~\cite{visweswariah2004first}.
What is new is reading it as the
gradient a softened contact approximates. The neighborhood-sampling remedy
of Section~\ref{sec:related}, with the belief as its neighborhood, estimates
\eqref{eq:smoothgrad} by Monte Carlo; the closed form is its draw-free
limit.

\subsection{The recipe: folding the maxima, differentiating the rest}
\label{sec:method}

The one-hot failure is one of \emph{where} the derivative is taken, not of the
derivative. A plan's cost is a computational graph of operations on the belief, whose inputs
are jointly Gaussian under Assumption~\ref{as:belief}; we build it from two
kinds of node. At a maximum we do not differentiate: we
compute the exact first two moments of the max (Clark), match a Gaussian to
them, and track every node's covariance against the shared map cells so that
later nodes receive their cross terms. At a smooth function we
differentiate: first-order propagation of the tracked moments is the delta
method, whose error is the Taylor remainder (Section~\ref{sec:delta}) and
which is free of Section~\ref{sec:onehot}'s pathology because a smooth node
has no winner to switch.

\begin{assumption}[Frozen rollout]\label{as:frozen}
Each candidate plan's pose sequence is computed once, on the mean map
$\bar{\mathbf{h}}$, and held fixed: the estimator carries the spread of
terrain under a fixed trajectory, not the spread of trajectories.
\end{assumption}

A sampling planner already uses the map this way: each candidate is
rolled out on the nominal model and its risk read along
it~\cite{fan2021step}; the estimator replaces the score, not the rollout.

\subsection{Step 1: the moments of one maximum}
\label{sec:fold}

Let $A$ and $B$ be jointly Gaussian with means $\mu_A, \mu_B$, variances
$\sigma_A^2, \sigma_B^2$ and covariance $\sigma_{AB}$. Clark's
identity~\cite{clark1961greatest} gives the exact first two moments of
$M = \max(A,B)$. With $a$ the width of their difference, as
in~\eqref{eq:diffwidth}, and the lead measured in its units,
\begin{equation}
  \alpha = \frac{\mu_A - \mu_B}{a},
  \label{eq:clark-alpha}
\end{equation}
\begin{align}
  \mathbb{E}[M]   &= \mu_A \Phi(\alpha) + \mu_B \Phi(-\alpha) + a\,\varphi(\alpha),
  \label{eq:clark-mean}\\
  \mathbb{E}[M^2] &= (\mu_A^2 + \sigma_A^2)\,\Phi(\alpha)
                   + (\mu_B^2 + \sigma_B^2)\,\Phi(-\alpha) \nonumber\\
                  &\quad + (\mu_A + \mu_B)\, a\, \varphi(\alpha).
  \label{eq:clark-m2}
\end{align}
$M$ is not Gaussian, but we \emph{moment-match} it to one and fold again.
This is the construction's one approximation, the one SSTA rests
on~\cite{visweswariah2004first}. We fold a footprint of
$K = |\mathcal{F}|$ candidates pairwise, in descending order of mean; at a
degenerate contest, $a = 0$, the fold keeps the leader's moments. The
recursion is order-dependent in principle, but not on the design traverses:
ascending, variance-ordered and random orders move the mean by under $0.1\%$
and the standard deviation (sd) by under $0.35\%$ at the median, with the
same aggregate deficit (post hoc).
We call the resulting estimator the \emph{Clark fold}. With
$\mu_A \geq \mu_B$, the excess of
$\mathbb{E}[M]$ over the larger mean is
$a\,\psi(\alpha) = a\,[\varphi(\alpha) - \alpha\,\Phi(-\alpha)]$, the
Jensen inflation of Section~\ref{sec:onehot} computed rather than ignored.

For any other jointly Gaussian quantity $Z$,
\begin{equation}
  \mathrm{Cov}(M, Z) \;=\; \Phi(\alpha)\,\mathrm{Cov}(A, Z)
                      \;+\; \Phi(-\alpha)\,\mathrm{Cov}(B, Z),
  \label{eq:clark-cov}
\end{equation}
exact for the \emph{true} maximum when $(A, B, Z)$ are jointly Gaussian
(Clark's, unchanged); carried forward as the matched Gaussian's covariance it
is part of the same moment-matching approximation, at one multiply per tracked
quantity per fold.
Every contact at every step is its own maximum and their footprints overlap
on shared cells, so each folded node is tracked against the shared universe
of raw map cells, SSTA's canonical first-order
form~\cite{visweswariah2004first}. A second pass replays each node's folds
against another node's universe vector and yields the node-to-node
covariance matrix $C$, which the attitude cost's quadratic form consumes
and a timing graph does not.

\subsection{Step 2: the delta method for the rest}
\label{sec:delta}

Past the maxima, collect the folded supports, one per contact per step,
into a vector $\mathbf{e}$, with means from step 1 and covariance $C$. Any
smooth $f$ then propagates by the delta method,
\begin{equation}
  \mathbb{E}[f(\mathbf{e})] \approx f(\mathbb{E}[\mathbf{e}]),
  \qquad
  \mathrm{Var}[f(\mathbf{e})] \approx \nabla f^{\top} C \, \nabla f ,
\end{equation}
with Taylor-remainder error, exact for an affine $f$. A quadratic
$f(\mathbf{e}) = \mathbf{e}^{\top}\!Q\,\mathbf{e}$ with symmetric $Q$ (ours
is $Q = G^{\top}G$, symmetric by construction) needs no expansion: with
$\mathbf{m} = \mathbb{E}[\mathbf{e}]$, the Gaussian quadratic-form
identities (the step that~\cite{tomita2023stochastic} performs for slope)
\begin{align}
  \mathbb{E}[f] &= \operatorname{tr}(Q C) + \mathbf{m}^{\top}\!Q\,\mathbf{m},
  \label{eq:quadform}\\
  \mathrm{Var}[f] &= 2\operatorname{tr}(Q C Q C)
    + 4\,\mathbf{m}^{\top}\!Q\,C Q\,\mathbf{m}
  \label{eq:quadvar}
\end{align}
replace it. Equation~\eqref{eq:quadform} is exact given $(\mathbf{m}, C)$,
and~\eqref{eq:quadvar} is exact for jointly Gaussian $\mathbf{e}$. While the
moment-matched supports are not strictly Gaussian, the swap test of
Section~\ref{sec:estimator} prices this gap at $0.4\%$.
The evaluated path cost (squared pitch and roll summed along the plan, which
are linear in the supports) takes exactly this quadratic form.

\subsection{Step 2b: closing the sd deficit}
\label{sec:estimator}

Step 1's moment matching, the chain's sole approximation
(Section~\ref{sec:fold}), is least accurate in the second moment, particularly
where candidate heights are nearly tied, the regime SSTA finds
hardest~\cite{sinha2007advances}.
Fold by fold, however, the matched Gaussian is faithful: measured against Monte Carlo on
each individual maximum, its median sd is within
$1.4\%$ of the truth in every contest-depth bin (Section~\ref{sec:gate}). The deficit
appears only once the folds are aggregated, the predicted sd falling a
median $6.1\%$ and $3.5\%$ short of the reference over a plan
(Section~\ref{sec:experiments}-C). A swap test locates it. On $40$ held-out
windows, \eqref{eq:quadvar} evaluated on the fold's joint Gaussian returns
$0.935$ of the reference sd and on the reference's own support moments
$0.996$; of that gap, the fold's marginals account for $5\%$ and its
cross-covariances for $86\%$. The deficit is the fold's own, but it exists
only in aggregation: each matched Gaussian is faithful on its own margin and
slightly too independent of its neighbors. We close it by calibration.

\begin{figure}[t]
  \centering
  \includegraphics[width=\columnwidth]{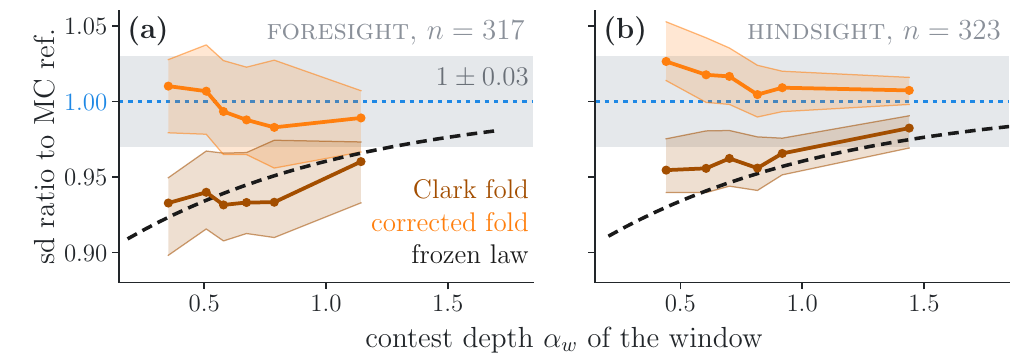}
  \caption{The standard deviation deficit and correction on held-out
  windows. These plots show the ratio of each estimator's standard deviation
  to the Monte Carlo reference, plotted against the window's contest depth
  $\alpha_w$. The data is visualized as medians and interquartile ribbons across
  six $\alpha_w$-quantile bins, with the fitted correction law $d(\alpha_w)$
  overlaid. \textbf{(a)} Foresight: the estimators and the deficit
  correction under the causal map available at decision time.
  \textbf{(b)} Hindsight: the same under the fully fused map. In (b) the
  uncorrected fold's medians sit above the frozen law by $0.004$ to $0.025$;
  in (a) they cross it.}
  \label{fig:deficit}
\end{figure}

Contest depth is measurable
from the belief alone, as the cost-weighted median $\alpha_w$ of Clark's
$\alpha$ over a plan's folds. On design data the ratio of a plan's matched sd
to its true sd, a window-level ratio and not a per-fold one, follows
$d(\alpha_w) = 1 - \beta \exp(-\gamma\,\alpha_w)$; the
\emph{corrected} fold, the estimator we propose, multiplies the predicted sd
by $1/d(\alpha_w)$. The form is a fit rather than a derivation, a
two-parameter family with $d \to 1$ as $\alpha_w$ grows, and
$(\beta, \gamma) = (0.110, 1.02)$ is its least-squares minimizer over $93$
design windows, each scored against Monte Carlo from its own belief and
frozen before any held-out pass (Section~\ref{sec:experiments},
Fig.~\ref{fig:deficit}). Both constants read directly: $\beta = 0.110$ is the
deficit at a fully contested plan, and $1/\gamma = 0.98$ is the contest
depth over which it decays. Re-deriving them at a different reference budget,
a pre-registered check, moves them to $(0.105, 0.971)$, inside the tolerance
frozen with the law.
Where contest
depth barely varies, $\gamma$ is not identifiable
(Section~\ref{sec:experiments}-C). A skew-normal
fold~\cite{chopra2006skew} would buy that back per fold,
which is not where the swap test puts our deficit; we have not tested it.

The computational graph of Section~\ref{sec:method} already computes the
fold's means and every $\alpha$, so the correction adds no operation to the
fold.

\subsection{The fold and its first-order limit}
\label{sec:gate}

First-order propagation is not a rival of the fold but its limit: as
$\alpha \to \infty$ in \eqref{eq:clark-mean}--\eqref{eq:clark-cov},
$\Phi(\alpha) \to 1$ and $\varphi(\alpha) \to 0$, and the fold returns the
leader's mean, variance and covariance row, the one-hot derivative
\eqref{eq:onehot}. What the fold adds at finite $\alpha$ is the
excess $a\,\psi(\alpha)$ of Section~\ref{sec:fold}, which falls from $0.399$
at $\alpha = 0$ to $0.0085$ at $\alpha = 2$; SSTA treats such dominated folds
as negligible~\cite{sinha2007advances}.
Post hoc, on the two design traverses, $68\%$ and $73\%$ of the $216{,}000$ pairwise folds
have $\alpha \geq 2$, and skipping them forgoes $0.63\%$ and $0.55\%$ of the
Jensen lift:
linearization suffices there, and the fold earns its accuracy in the
contested third that remains. We nonetheless deploy the fold everywhere,
since a settled fold costs the same handful of multiplies as a skipped one.

The variance has no such limit at finite $\alpha$. Subtracting the square
of~\eqref{eq:clark-mean} from~\eqref{eq:clark-m2} at a tie, $\alpha = 0$,
gives
\begin{equation}
  \mathrm{Var}[M] \;=\; \tfrac{1}{2}\big(\sigma_A^2 + \sigma_B^2\big)
  \;-\; \frac{a^2}{2\pi},
  \label{eq:tie-var}
\end{equation}
where linearization keeps $\sigma_A^2$. For equal variances the two agree
only as $a \to 0$, the cells then being one variable; two uncorrelated cells
have $a^2 = 2\sigma^2$ and a true variance of
$\sigma^2(1 - 1/\pi) = 0.68\,\sigma^2$, so linearization overstates the
standard deviation by $21\%$. Linearization's variance is thus exact in two
regimes only, settled folds ($\alpha \geq 2$) and coincident cells
($a \ll \sigma$). Between them it overstates: against Monte Carlo on each
maximum, over the $21{,}144$ maxima of the design foresight windows its
standard deviation is $1.28$ times the true one at $\alpha < 0.25$, $1.19$
pooled and $1.000$ once $\alpha \geq 4$, understating in only $7.4\%$ of
folds, while the fold's own median stays within $1.4\%$ in every bin.

\subsection{Step 3: the ranking rule, and what it costs}
\label{sec:cost}

Plans are ranked by a Gaussian CVaR at level $q$, built from the two moments
of the plan cost $\cost = f(\mathbf{e})$,
\begin{equation}
  \hat{\cost} \;=\; \mathbb{E}[\cost] \;+\; \lambda_q \sqrt{\mathrm{Var}[\cost]},
  \qquad
  \lambda_q = \frac{\varphi(\Phi^{-1}(q))}{1 - q},
  \label{eq:cvar}
\end{equation}
at $q = 0.90$. This is the Gaussian CVaR that STEP uses as its one-step
risk metric, compounded along the trajectory (Eqs.~4 and~15
of~\cite{fan2021step}). The difference is not the risk functional but the moments
fed to it.

On an RTX 3090, on the pre-registration's synthetic benchmark scene ($P = 256$
plans, $T = 40$ steps), the fold's two moments cost $5.5~\mu$s per plan:
$5.8\times$ one deterministic cost evaluation and $0.4\times$ the compute of
$32$-draw sampling, whose moment error on the held-out record is
$3.3$ to $11\times$ larger (Table~\ref{tab:moments}). Equal compute is
therefore about $13$ draws, so \textsc{mc-2} is cheaper than the fold rather
than matched to it, and \textsc{mc-32} is dearer. On the held-out record, scaling \textsc{mc-32}'s error by $1/\sqrt{D}$ puts the
break-even at about $1{,}300$ and $3{,}900$ draws for the mean, where the
sampler costs $100$ to $300\times$ the fold.

\section{Experiments}
\label{sec:experiments}

To validate the corrected fold, we evaluate two pre-registered claims
against a Monte Carlo (MC) reference:

\begin{claim}[Moment Recovery]\label{cl:estimator}
  At a contested contact, standard linearization recovers neither moment of
  the contact height. Over a path, the fold recovers the mean of the cost,
  and the corrected fold (its standard deviation calibrated on design data)
  recovers the standard deviation.
\end{claim}

\begin{claim}[Decision Accuracy]\label{cl:decision}
  The corrected fold achieves the highest accuracy for
  $\mathrm{CVaR}_{0.9}$ among all tested methods, surpassing both existing
  analytical approximations and sampling approaches.
\end{claim}

We benchmark all methods against an MC reference of at least $20{,}000$
terrains per window. By drawing these from the same belief map used by the
estimators, we isolate propagation accuracy from mapping error. The
reference $\mathrm{CVaR}_{0.9}$ is the empirical mean of the worst $10\%$ of
sampled costs. This differs from the Gaussian form~\eqref{eq:cvar} by under
$1\%$ at the median, the error scale any Gaussian-tail estimator carries.
All comparative success criteria were pre-registered prior to the single
held-out evaluation pass. Claim~\ref{cl:estimator} is deliberately scoped by
level: standard linearization's per-contact failure does not carry over to
the aggregate standard deviation over a path, which
Section~\ref{sec:moments} measures as unbiased.
Artifacts, scoring code and the pre-registration will be released; the
frozen pass's seeds were unrecorded, so its reference moments are released
directly and later studies redraw with recorded seeds.

\subsection{Common Protocol}

We employ a quasi-static reduction of a differentiable contact
engine~\cite{kucera2026ostrich} to model a three-wheeled skid-steer
platform.
Each wheel's contact height is evaluated as the maximum~\eqref{eq:envelope}
over $K = 5$--$7$ candidate cells at $0.10$~m resolution, offset by the
measured tread cylinder.
The quasi-static solver maps these three supports to pitch $\theta_t$ and
roll $\phi_t$ via affine rows. This defines the path cost as the cumulative
attitude functional $\cost = \sum_{t} ( \theta_t^{2} + \phi_t^{2} )$.
Because we score the executed trajectory, Assumption~\ref{as:frozen} holds
by construction.

The unit of prediction is a ``window'': a contiguous $4$~m trajectory
segment resampled into $0.10$~m steps. For every window that passes our
geometry-only quality flags (observed fraction and belief variance), each
method must predict the mean $\mathbb{E}[\cost]$, standard deviation
$\mathrm{sd}[\cost]$, and $\mathrm{CVaR}_{0.9}$ of the path cost. We score
relative error for the two moments and absolute error for the risk, the
latter written out in Table~\ref{tab:cvar}.

\subsection{Dataset and Experimental Setup}
\label{sec:setup}

We evaluate our method on BASEPROD~\cite{gerdes2024baseprod}, a semi-desert
rover dataset. Elevation maps are constructed from raw depth
streams~\cite{fankhauser2018probabilistic}. Because the
mapper outputs only a marginal variance per cell, we construct a spatial
covariance model (incorporating independent measurement error, globally
shared uncertainty, and localization drift) that exactly recovers these
marginals. This covariance is a derived model, not a direct measurement.
Its composition is not a free hand, however: over the $1.98$ million observed
held-out cells the independent term carries $96.7\%$ of the variance and the
shared term $3.3\%$ and the localization term under $10^{-5}$.
Neighboring
cells are thus effectively independent and a shared offset cancels from the
contest scale $a$, so the contest depths below follow from the published
marginals, not from a correlation length of ours.

Our primary evaluation condition, foresight, halts map fusion before the
rover enters the window to simulate the causal map available at decision
time. As a control for map quality, hindsight fuses the entire traverse. Of
the $24$ available traverses, two serve as design data to fit the standard
deviation correction (Section~\ref{sec:estimator}). A pre-registered
kinematic filter excluded one traverse,
and three lacked windows passing our geometric quality flags. This leaves
$18$ held-out traverses, yielding $317$ foresight and $323$ hindsight
windows.

Our proposed estimator, the corrected fold (\textsc{clark-corr}), requires
design data to scale the fold's standard deviation by $1/d(\alpha_w)$, where
$d = 1 - 0.110\,e^{-1.02\,\alpha_w}$ was fitted on the design traverses and
frozen. Pre-registered baselines include standard linearization (\textsc{fosm}), the
pure moment-matched fold (\textsc{clark}), that fold with its off-diagonal
covariance removed (\textsc{clark-diag}), Monte Carlo with $2$ and $32$
samples (\textsc{mc-2} and \textsc{mc-32}), the
mean map, and a STEP
analog~\cite{fan2021step} (\textsc{step-form}) that upper-bounds the
trajectory's $\mathrm{CVaR}$ by construction. Finally, we include two post
hoc baselines: a softened-contact arm (\textsc{lse}) using a per-node
temperature $\tau = a/1.702$ to match the logistic weight to the Gaussian
win probability $\Phi(\Delta/a)$, and a scaled unscented
transform~\cite{julier2002scaled} (\textsc{ut}) with $2N + 1$ sigma points,
swept over five spreads.

\begin{figure}[t]
  \centering
  \includegraphics[width=0.95\columnwidth]{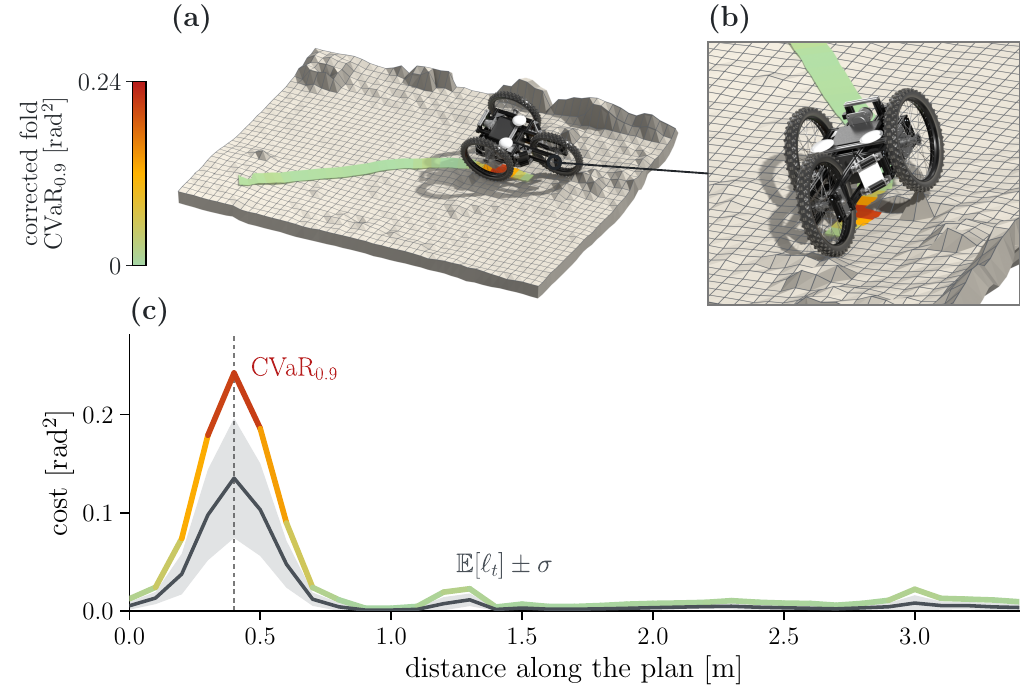}
  \caption{The corrected fold on a BASEPROD elevation map. \textbf{(a)} The
  belief mean across a $4$~m foresight window of a design traverse, rendered
  orthographically at true vertical scale. The path is the window's own track
  (the $35$ of $39$ steps that pass the quality flags), colored by the
  corrected fold's $\mathrm{CVaR}_{0.9}$ at each step. A $3$D model of the
  robot is placed where this risk peaks, its wheels seated on the support
  heights the fold returns there: the rear wheel's cylinder element evaluates
  $K = 7$ cells, where the top two candidates are nearly tied, $\alpha =
  0.01$ standard deviations of their difference apart. The vehicle's
  $20^\circ$ roll, derived from these three supports, is computed but not
  drawn. \textbf{(b)} A $1.96$~m wide rear view of the contested rear
  contact. It reveals the ground beneath that wheel, which the chassis and
  the front wheels obscure in panel (a). The ring in (a) marks this contact.
  \textbf{(c)} The per-step mean $\mathbb{E}[\cost_t]$ alongside its
  $\pm 1\sigma$ band, the ribbon colored by the $\mathrm{CVaR}_{0.9}$ risk.
  The risk spikes to $0.24$~rad$^2$ at the contested step, against a median
  of $0.0094$ along the plan.}
  \label{fig:rover}
\end{figure}

\begin{table}[t]
\caption{Claim~\ref{cl:estimator}: median relative error against the MC
reference, \textsc{foresight}/\textsc{hindsight} ($317$ / $323$ windows).
Bold marks the best in each column; the two Clark arms share one mean.}
\label{tab:moments}
\centering\footnotesize
\begin{tabular}{@{}lrr@{}}
\toprule
 & med.\ rel.\ err.\ $\mathbb{E}[\cost]$ & med.\ rel.\ err.\ $\mathrm{sd}[\cost]$ \\
\midrule
\textsc{clark-corr} (ours) & $\mathbf{1.9\!\cdot\!10^{-3}}$ / $\mathbf{8.5\!\cdot\!10^{-4}}$ & $\mathbf{0.028}$ / $\mathbf{0.018}$ \\
\textsc{clark}   & $\mathbf{1.9\!\cdot\!10^{-3}}$ / $\mathbf{8.5\!\cdot\!10^{-4}}$ & $0.061$ / $0.035$ \\
\textsc{fosm}    & $2.3\!\cdot\!10^{-2}$ / $1.5\!\cdot\!10^{-2}$ & $0.038$ / $0.023$ \\
\textsc{mc-32}   & $1.2\!\cdot\!10^{-2}$ / $9.4\!\cdot\!10^{-3}$ & $0.093$ / $0.097$ \\
\bottomrule
\end{tabular}
\end{table}

\subsection{Claim~\ref{cl:estimator}: Mean and Standard Deviation}
\label{sec:moments}

The corrected fold achieves a median relative
error for the mean path cost of $0.19\%$ (foresight) and $0.085\%$
(hindsight), compared to $2.3\%$ and $1.5\%$ for standard linearization.
Evaluated per
window, the fold yields a smaller mean error than linearization in
$301$ of $317$ foresight and $311$ of $323$ hindsight windows, and maintains
a majority across all $18$ traverses (exact sign test
$p = 7.6\times10^{-6}$).

The pure fold understates the standard deviation, yielding median ratios to
the MC reference of $0.939$ and $0.965$. Applying the corrected fold's
two-parameter law (fitted on design data) shifts these ratios to $0.996$ and
$1.014$ (Fig.~\ref{fig:deficit}), cutting the median relative error from
$6.1\%$ to $2.8\%$ in foresight, and $3.5\%$ to $1.8\%$ in hindsight. Post
hoc, a single scalar fits the held-out record just as well ($3.0\%$ and
$1.8\%$), indicating the identifiability limit of contest depth is reached
on this dataset. Linearization's standard deviation carries no such bias
(ratios $1.000$ and $0.997$), placing its median errors of $3.8\%$ and
$2.3\%$ between the pure and corrected folds. On this dataset
linearization's failure is confined to the mean, but that
is an aggregate offset rather than a per-fold property: on each
individual maximum linearization's sd runs $1.19$ times the truth in the
median and $1.28$ times at the most contested quarter
(Section~\ref{sec:gate}). The cancellation is nonetheless robust to the
mixture it averages over: across the covariance sweep below, which lifts the
cost-weighted contest depth from $0.59$ to $1.28$, linearization's sd ratio
stays between $0.994$ and $1.002$.

Because that mixture is set by a covariance we model rather than measure, we
test how much of this result depends on it. Holding every published marginal
fixed, we move a fraction $f$ of each cell's residual variance into a term
correlated over $0.15$~m and re-score $77$ held-out foresight windows
($6{,}000$ draws; post hoc). Correlating the cells does settle the contests,
the cost-weighted contest depth rising from $0.59$ to $1.28$, and
linearization's median error of the mean duly falls from $2.3\%$ to
$0.86\%$. The fold's advantage narrows but holds throughout: its mean error
stays $4.5$ to $13$ times below linearization's, winning $91$ to $97\%$ of
windows, and the corrected fold's $\mathrm{CVaR}$ error stays $1.5$ to $2.9$
times below, winning $71$ to $79\%$.

In a zero-shot post hoc transfer to $110$ Oxford Spires
windows~\cite{tao2025spires}, outside the pre-registered
$\sigma \leq 0.10$~m flag, the fold's mean transferred: a lower error in
$108$ of them ($1.1$--$2.7\%$ against linearization's $6.2$--$8.2\%$). The
risk number did not. On roof-aliased windows the frozen law overshot and the
reference cost is far from Gaussian, so the corrected fold's
$\mathrm{CVaR}$ error exceeds linearization's at two of the three sites.
Outside the registered regime the ranking rule fails while the moment
propagation holds.

The corrected fold also tracks both reference moments more accurately than a
$32$-sample Monte Carlo baseline in $231$ and $255$ windows, a majority in
all $18$ traverses.

\begin{table}[t]
\caption{Claim~\ref{cl:decision}:
$|\mathrm{CVaR}_{0.9}(\mathrm{method}) - \mathrm{CVaR}_{0.9}(\mathrm{MC})|$
(rad$^2$), median and 95th percentile, \textsc{foresight} (F) and
\textsc{hindsight} (H), on the $317$ / $323$ held-out windows (reference
median $0.50$ / $0.63$); the \textsc{lse} and \textsc{ut} rows use all
$363$ / $368$ windows and a re-drawn reference, where \textsc{clark-corr} is
$0.0045$ / $0.0019$.}
\label{tab:cvar}
\centering\footnotesize
\begin{tabular}{@{}lrrrr@{}}
\toprule
method & F med. & F p95 & H med. & H p95 \\
\midrule
\textsc{clark-corr} (ours) & $\mathbf{0.0045}$ & $\mathbf{0.124}$ & $\mathbf{0.0022}$ & $\mathbf{0.152}$ \\
\textsc{clark}      & $0.0088$ & $0.234$ & $0.0054$ & $0.257$ \\
\textsc{mc-32}      & $0.0085$ & $0.322$ & $0.0068$ & $0.484$ \\
\textsc{fosm}       & $0.012$ & $0.823$ & $0.0074$ & $0.778$ \\
\textsc{lse}          & $0.015$ & $0.179$ & $0.0090$ & $0.196$ \\
\textsc{ut}           & $0.027$ & $1.216$ & $0.016$ & $1.028$ \\
\textsc{mc-2}       & $0.038$ & $1.698$ & $0.036$ & $1.722$ \\
mean map            & $0.077$ & $3.274$ & $0.069$ & $3.532$ \\
\textsc{step-form}  & $0.352$ & $3.564$ & $0.377$ & $3.651$ \\
\bottomrule
\end{tabular}
\end{table}

\subsection{Claim~\ref{cl:decision}: Risk Estimate}

The softened-contact baseline (\textsc{lse}) measures
Section~\ref{sec:softening}'s argument that no temperature recovers the win
probability. The belief-scaled softmax partially recovers the mean, but understates the
standard deviation by $10\%$ and $6.5\%$ and so returns a higher
$\mathrm{CVaR}$ error than plain linearization. A scaled unscented
transform~\cite{julier2002scaled} fails from the other side. Its sigma points
sit at a fixed multiple of $\sigma$, so at the standard spread a different
cell wins at each and the predicted standard deviation is $16.8$ and $9.5$
times the reference, while shrinking the spread returns them to the winner's
region and recovers linearization; the best of the five spreads we swept,
$0.03$, still leaves the $\mathrm{CVaR}$ error six to eight times ours.
Neither a temperature nor a sigma-point scale is a route to the second
moment.

The corrected fold achieves the lowest median $\mathrm{CVaR}_{0.9}$ error of
all tested methods, $0.0045$ and $0.0022$ (Table~\ref{tab:cvar}): $2.7$ to
$3.4$ times below linearization at the median, $5.1$ to $6.6$ times at the
95\textsuperscript{th} percentile, and the lower error in $16$ and $18$ of
the $18$ traverses.
Tracking cross-node
covariance is essential: dropping the fold's off-diagonal terms
(\textsc{clark-diag}) multiplies the pure fold's median $\mathrm{CVaR}$ error by
$3.4$ and $4.6$, to $0.030$ and $0.025$.

The pure fold falls between the corrected fold and linearization ($0.0088$
and $0.0054$); since the two share a mean, the standard deviation deficit
(Section~\ref{sec:estimator}) is what degrades the risk metric.

At more conservative quantiles our error stays below linearization's, but at
$q = 0.99$ our $2.6\%$ essentially reaches the $2.55\%$ floor that any
estimator feeding exact moments into a Gaussian tail must pay: there the
tail of~\eqref{eq:cvar}, not the propagation, is the bottleneck.

\subsection{Path Selection}

From the start of each window we generate $32$ candidate paths of $3$~s,
each scored as a fixed trajectory (Assumption~\ref{as:frozen}) and ranked by
predicted $\mathrm{CVaR}_{0.9}$ against a reference best path drawn with an
$8{,}000$-sample budget. After discarding candidates exceeding $0.35$~rad of
pitch or roll on the mean map, $284$ foresight and $289$ hindsight decisions
remain.

Median regret (the difference in reference $\mathrm{CVaR}_{0.9}$ between the
chosen and best paths) is zero for all evaluated methods. The corrected fold selects the
reference best path in $98\%$ of decisions under both conditions, as does
the uncorrected fold. This outperforms standard linearization
($96\%$ and $93\%$), $32$-sample Monte Carlo ($90\%$ and $89\%$), and the
mean map ($90\%$ and $88\%$). Evaluated via an exact McNemar test on
discordant top-$1$ selections, our margin over linearization is $10$ to $3$
under foresight ($p = 0.09$, not significant) and $13$ to $0$ under
hindsight ($p = 2.4\times10^{-4}$). The corrected fold is not statistically
separated from the pure fold on this metric. Note that
the two conditions share the same traverses, representing two readings of
one record rather than independent tests.

To determine whether this accuracy alters discrete safety verdicts, we
evaluated post hoc the threshold crossings at the reference
$\mathrm{CVaR}_{0.9}$ quartiles and $90$th percentile. The corrected fold
and standard linearization disagree with the reference on $0$ to $1.6\%$ of
windows, the mean map on $1.9$ to $7.6\%$ (calling unsafe windows safe) and
the STEP analog on $1.3$ to $17.7\%$ (rejecting safe ones). On this terrain the verdicts separate the two folds from the mean
map and STEP but not from linearization, over which the advantage
is in the tail (Table~\ref{tab:cvar}).

\section{Conclusion}
\label{sec:discussion}
\label{sec:limitations}

Path costs beneath a rigid wheel are governed by a maximum over uncertain
terrain heights. Standard linearization differentiates at the cell that wins
on the mean map, losing both moments in the process, and no softening of the
contact can recover them. We address this by folding the maximum in closed
form.
The corrected Clark fold recursively evaluates contact cells while tracking
covariance. While its standard deviation is accurate fold by fold, the
accuracy it loses in aggregation over a plan is restored by a contest-depth
calibration fitted on two design traverses. Without this calibration, the
fold's standard deviation is less accurate than linearization's, which
happens to remain unbiased in aggregate on this record despite failing fold
by fold.
Executing in just $5.5~\mu$s per plan on a GPU, the corrected fold reduces
median $\mathrm{CVaR}_{0.9}$ error by a factor of $2.7$ to $3.4$ compared to
standard linearization. This advantage is earned at contested contacts;
where a single cell dominates its footprint, the fold and linearization
behave identically.

Our evaluation grades propagation against the map's internal
belief rather than physical ground truth. A comparison against a drone
survey reveals that the raw map severely understates physical elevation
errors: the fold's $\pm 1\sigma$ band covers surveyed supports at just $20$
to $21\%$ of nodes, versus a nominal $68\%$. This indicates that mapper
miscalibration outweighs propagation error on this record.
Furthermore, our contests are at their widest because the belief carries
almost no spatial correlation between cells; introducing correlation narrows
the gap to linearization without closing it
(Section~\ref{sec:experiments}-C). Our estimator also assumes a fixed
trajectory and a Gaussian tail, the latter breaking down in aliased
environments. Finally, folding each wheel's contact independently is exact
only when the supports fully determine the pose (as in the three-point
stance evaluated here), and the delta-method step is not licensed
for non-smooth functionals like clearance violations.

Future work includes carrying the folded supports into a coupled contact
solve, which would extend the recipe to platforms with redundant contacts
and to dynamic costs, and developing probabilistic mappers whose beliefs are
calibrated directly against realized contact supports.

\bibliographystyle{IEEEtran}
{\scriptsize\linespread{0.90}\selectfont
\bibliography{references}}

\begin{thebibliography}{10}
\providecommand{\url}[1]{#1}
\csname url@samestyle\endcsname
\providecommand{\newblock}{\relax}
\providecommand{\bibinfo}[2]{#2}
\providecommand{\BIBentrySTDinterwordspacing}{\spaceskip=0pt\relax}
\providecommand{\BIBentryALTinterwordstretchfactor}{4}
\providecommand{\BIBentryALTinterwordspacing}{\spaceskip=\fontdimen2\font plus
\BIBentryALTinterwordstretchfactor\fontdimen3\font minus
  \fontdimen4\font\relax}
\providecommand{\BIBforeignlanguage}[2]{{%
\expandafter\ifx\csname l@#1\endcsname\relax
\typeout{** WARNING: IEEEtran.bst: No hyphenation pattern has been}%
\typeout{** loaded for the language `#1'. Using the pattern for}%
\typeout{** the default language instead.}%
\else
\language=\csname l@#1\endcsname
\fi
#2}}
\providecommand{\BIBdecl}{\relax}
\BIBdecl

\bibitem{fankhauser2018probabilistic}
P.~Fankhauser, M.~Bloesch, and M.~Hutter, ``Probabilistic terrain mapping for
  mobile robots with uncertain localization,'' \emph{IEEE Robot. Autom. Lett.},
  vol.~3, no.~4, pp. 3019--3026, 2018.

\bibitem{fan2021step}
D.~D. Fan, K.~Otsu, Y.~Kubo, A.~Dixit, J.~Burdick, and A.-A. Agha-Mohammadi,
  ``{STEP}: Stochastic traversability evaluation and planning for risk-aware
  off-road navigation,'' in \emph{Robotics: Science and Systems (RSS)}, 2021.

\bibitem{clark1961greatest}
C.~E. Clark, ``The greatest of a finite set of random variables,'' \emph{Oper.
  Res.}, vol.~9, no.~2, pp. 145--162, 1961.

\bibitem{gerdes2024baseprod}
L.~Gerdes, T.~Wiese, R.~Castilla~Arquillo, L.~Bielenberg, M.~Azkarate,
  H.~Leblond, F.~Wilting, J.~Ortega~Cort{\'e}s, A.~Bernal, S.~Palanco, and
  C.~P{\'e}rez~del Pulgar, ``{BASEPROD}: The {Bardenas} semi-desert planetary
  rover dataset,'' \emph{Scientific Data}, vol.~11, p. 1054, 2024.

\bibitem{otsu2020ace}
K.~Otsu, G.~Matheron, S.~Ghosh, O.~Toupet, and M.~Ono, ``Fast approximate
  clearance evaluation for rovers with articulated suspension systems,''
  \emph{J. Field Robot.}, vol.~37, no.~5, pp. 768--785, 2020.

\bibitem{ghosh2018pace}
S.~Ghosh, K.~Otsu, and M.~Ono, ``Probabilistic kinematic state estimation for
  motion planning of planetary rovers,'' in \emph{IEEE/RSJ Int. Conf. Intell.
  Robots Syst. (IROS)}, 2018, pp. 5148--5154.

\bibitem{tomita2023stochastic}
K.~Tomita and K.~Ho, ``Stochastic hazard detection for landing under
  topographic uncertainty,'' in \emph{AAS/AIAA Space Flight Mechanics Meeting},
  2023, {AAS} 23-391, arXiv:2305.04249.

\bibitem{raja2025proterrain}
G.~Raja, R.~Agishev, M.~Pr{\'a}gr, J.~Pajarinen, K.~Zimmermann, A.~K. Singh,
  and R.~Ghabcheloo, ``{ProTerrain}: Probabilistic physics-informed rough
  terrain world modeling,'' \emph{arXiv preprint arXiv:2510.19364}, 2025.

\bibitem{newbury2024review}
R.~Newbury, J.~Collins, K.~He, J.~Pan, I.~Posner, D.~Howard, and A.~Cosgun, ``A
  review of differentiable simulators,'' \emph{IEEE Access}, vol.~12, pp.
  97\,581--97\,604, 2024.

\bibitem{agishev2024monoforce}
R.~Agishev, K.~Zimmermann, V.~Kubelka, M.~Pecka, and T.~Svoboda, ``{MonoForce}:
  Self-supervised learning of physics-informed model for predicting
  robot-terrain interaction,'' in \emph{IEEE/RSJ Int. Conf. Intell. Robots
  Syst. (IROS)}, 2024.

\bibitem{kucera2026ostrich}
A.~Ku{\v{c}}era and K.~Zimmermann, ``Ostrich: Taking large strides through
  stiff contact in differentiable dynamics,'' 2026, arXiv:2609.08800.

\bibitem{zhong2022correct}
Y.~D. Zhong, J.~Han, and G.~O. Brikis, ``Differentiable physics simulations
  with contacts: Do they have correct gradients w.r.t. position, velocity and
  control?'' \emph{arXiv preprint arXiv:2207.05060}, 2022.

\bibitem{paulus2025diffmjx}
A.~Paulus, A.~R. Geist, P.~Schumacher, V.~Musil, S.~Rappenecker, and
  G.~Martius, ``Differentiable simulation of hard contacts with soft gradients
  for learning and control,'' in \emph{Int. Conf. Learn. Represent. (ICLR)},
  2026.

\bibitem{lelidec2022randomized}
Q.~Le~Lidec, L.~Montaut, C.~Schmid, I.~Laptev, and J.~Carpentier, ``Augmenting
  differentiable physics with randomized smoothing,'' \emph{arXiv preprint
  arXiv:2206.11884}, 2022.

\bibitem{suh2022bundled}
H.~J.~T. Suh, T.~Pang, and R.~Tedrake, ``Bundled gradients through contact via
  randomized smoothing,'' \emph{IEEE Robot. Autom. Lett.}, vol.~7, no.~2, pp.
  4000--4007, 2022.

\bibitem{visweswariah2004first}
C.~Visweswariah, K.~Ravindran, K.~Kalafala, S.~G. Walker, and S.~Narayan,
  ``First-order incremental block-based statistical timing analysis,'' in
  \emph{Proc. Des. Autom. Conf. (DAC)}, 2004, pp. 331--336.

\bibitem{chopra2006skew}
K.~Chopra, B.~Zhai, D.~Blaauw, and D.~Sylvester, ``A new statistical max
  operation for propagating skewness in statistical timing analysis,'' in
  \emph{IEEE/ACM Int. Conf. Computer-Aided Design (ICCAD)}, 2006, pp. 237--243.

\bibitem{andrews2022contact}
S.~Andrews, K.~Erleben, and Z.~Ferguson, ``Contact and friction simulation for
  computer graphics,'' in \emph{ACM SIGGRAPH 2022 Courses}, 2022.

\bibitem{lozanoperez1983spatial}
T.~Lozano-P{\'e}rez, ``Spatial planning: A configuration space approach,''
  \emph{IEEE Trans. Comput.}, vol. C-32, no.~2, pp. 108--120, 1983.

\bibitem{sorensen2022mcfadden}
J.~R.-V. S{\o}rensen and M.~Fosgerau, ``How {McFadden} met {Rockafellar} and
  learned to do more with less,'' \emph{J. Math. Econ.}, vol. 100, p. 102629,
  2022.

\bibitem{sinha2007advances}
D.~Sinha, H.~Zhou, and N.~V. Shenoy, ``Advances in computation of the maximum
  of a set of {Gaussian} random variables,'' \emph{IEEE Trans. Comput.-Aided
  Des. Integr. Circuits Syst.}, vol.~26, no.~8, pp. 1522--1533, 2007.

\bibitem{julier2002scaled}
S.~J. Julier, ``The scaled unscented transformation,'' in \emph{Proc. American
  Control Conf. (ACC)}, vol.~6, 2002, pp. 4555--4559.

\bibitem{tao2025spires}
Y.~Tao, M.~{\'A}. Mu{\~n}oz-Ba{\~n}{\'o}n, L.~Zhang, J.~Wang, L.~F.~T. Fu, and
  M.~Fallon, ``The {Oxford Spires} dataset: Benchmarking large-scale
  {LiDAR}-visual localisation, reconstruction and radiance field methods,''
  \emph{Int. J. Robot. Res.}, 2025.

\end{thebibliography}

\end{document}